\documentclass{INTERSPEECH2023}
\usepackage{multirow}
\usepackage{subfigure}
\usepackage{amssymb}
\usepackage{cancel}

\interspeechcameraready

\title{Bayes Risk Transducer: Transducer with Controllable Alignment Prediction}
\name{Jinchuan Tian$^{1,3}$, Jianwei Yu$^{1,2}$, Hangting Chen$^1$, Brian Yan$^3$, Chao Weng$^{1,2}$, \\ Dong Yu$^1$, Shinji Watanabe$^3$}
\address{
  $^1$Tencent AI LAB, $^2$Tencent ASR Oteam, \\
  $^3$Language Technologies Institute, Carnegie Mellon University}
\email{tomasyu@tencent.com, swatanab@andrew.cmu.edu}

\begin{document}

\maketitle
 
\begin{abstract}
Automatic speech recognition (ASR) based on transducers is widely used. In training, a transducer maximizes the summed posteriors of all paths. The path with the highest posterior is commonly defined as the predicted alignment between the speech and the transcription. While the vanilla transducer does not have a prior preference for any of the valid paths, this work intends to enforce the preferred paths and achieve controllable alignment prediction. Specifically, this work proposes Bayes Risk Transducer (BRT), which uses a Bayes risk function to set lower risk values to the preferred paths so that the predicted alignment is more likely to satisfy specific desired properties. We further demonstrate that these predicted alignments with intentionally designed properties can provide practical advantages over the vanilla transducer.  Experimentally, the proposed BRT saves inference cost by up to 46\% for non-streaming ASR and reduces overall system latency by 41\% for streaming ASR.\footnote{We gratefully acknowledge the support of NVIDIA Corporation with the donation of the GPUs used for this research.}
\footnote{Code available: https://github.com/espnet/espnet}

\end{abstract}

\section{Introduction}
Automatic speech recognition (ASR) based on transducers \cite{rnnt} is one of the most popular frameworks \cite{survey1, survey2, survey3}. 
In the past few years, a series of approaches have been proposed as extensions of the transducer with the goals of optimizing its recognition accuracy \cite{rna, two-pass}, language model integration \cite{hat}, flexibility \cite{gtct}, decoding efficiency \cite{mbt} and simplicity \cite{stateless, mono}, memory efficiency during training \cite{prune_rnnt, asru_memory, ney} and overall system latency during streaming decoding \cite{fastemit, delayrnnt, slt_rnnt,stream_shinji, selfalign, google_emit}.
During training, the vanilla transducer, as well as its extensions \cite{mbt, prune_rnnt, asru_memory, fastemit, delayrnnt, slt_rnnt, google_emit}, maximizes the summed posterior of all potential aligning sequences (a.k.a, \textit{paths}) between the speech and the transcription. 
In particular, these extensions achieve their goals by manipulating the transducer paths, such as allowing multi-frame big skips \cite{mbt}, pruning paths with minor posteriors with \cite{asru_memory, slt_rnnt, google_emit} or without \cite{prune_rnnt} reference alignment labels, discouraging blank emissions \cite{fastemit} and encouraging non-blank emissions \cite{delayrnnt}. This work provides another extension of the transducer, which also conducts manipulation over paths. Specifically, as a follow-up of the previous work \cite{brctc} which attempts to achieve controllable alignment prediction in CTC criterion \cite{ctc}, this work extends this controllability to the transducer model by taking its distinctive forward-backward process into consideration.

The alignment prediction of the transducer is commonly defined as the path with the highest posterior.
In vanilla transducer formulation, there is no prior preference among the paths since predicting each valid path will yield the correct textual transcription. Currently, the alignment selection among the paths (i.e., which path will become the predicted alignment) can hardly be affected by human intervention during training.
To achieve controllable alignment prediction in the transducer is exactly to intentionally choose paths with specific desired properties as the alignment prediction.
With this motivation, this work proposes an extension of the transducer called Bayes Risk Transducer (BRT), which adopts a Bayes risk function to intentionally enforce a preference for paths with the desired properties, so that the predicted alignments are more likely to be characterized by these properties.
Particularly, the original forward-backward algorithm of the transducer is revised into a divide-and-conquer approach: all paths are firstly divided into multiple exclusive groups and the groups with more favored properties are enforced by receiving lower risk values than the others.

This work further demonstrates that BRT with controllable alignment prediction has practical advantages over vanilla transducers. By designing various Bayes risk functions, we can obtain alignment predictions with desired properties that are specific to different task setups, which subsequently helps to offer novel solutions for two practical challenges in ASR: inference cost for non-streaming ASR and overall system latency for streaming ASR.
In the non-streaming setup, a Bayes risk function is designed to enforce the paths that emit the last non-blank predictions earlier. As a benefit, the last non-blank prediction occurs at an early time stamp so the inference cost can be reduced by terminating the decoding loop early without exploring all frames. 
In the streaming setup, another Bayes risk function is designed to encourage early emissions for all non-blank tokens. Thus, the model emits before waiting too long context and the latency for each non-blank token is reduced.
Experimentally, the former case accelerates non-streaming inference by up to 46\% and the latter case reduces the overall system latency of streaming ASR system by 41\%.
\vspace{-8pt}

\section{Bayes Risk Transducer}
\setlength{\belowcaptionskip}{-0.7cm}
\setlength{\abovecaptionskip}{-0.1cm}
\begin{figure*}[t]
\centering
\subfigure{
\begin{minipage}[t]{0.305\linewidth}
\centering
\includegraphics[width=\linewidth]{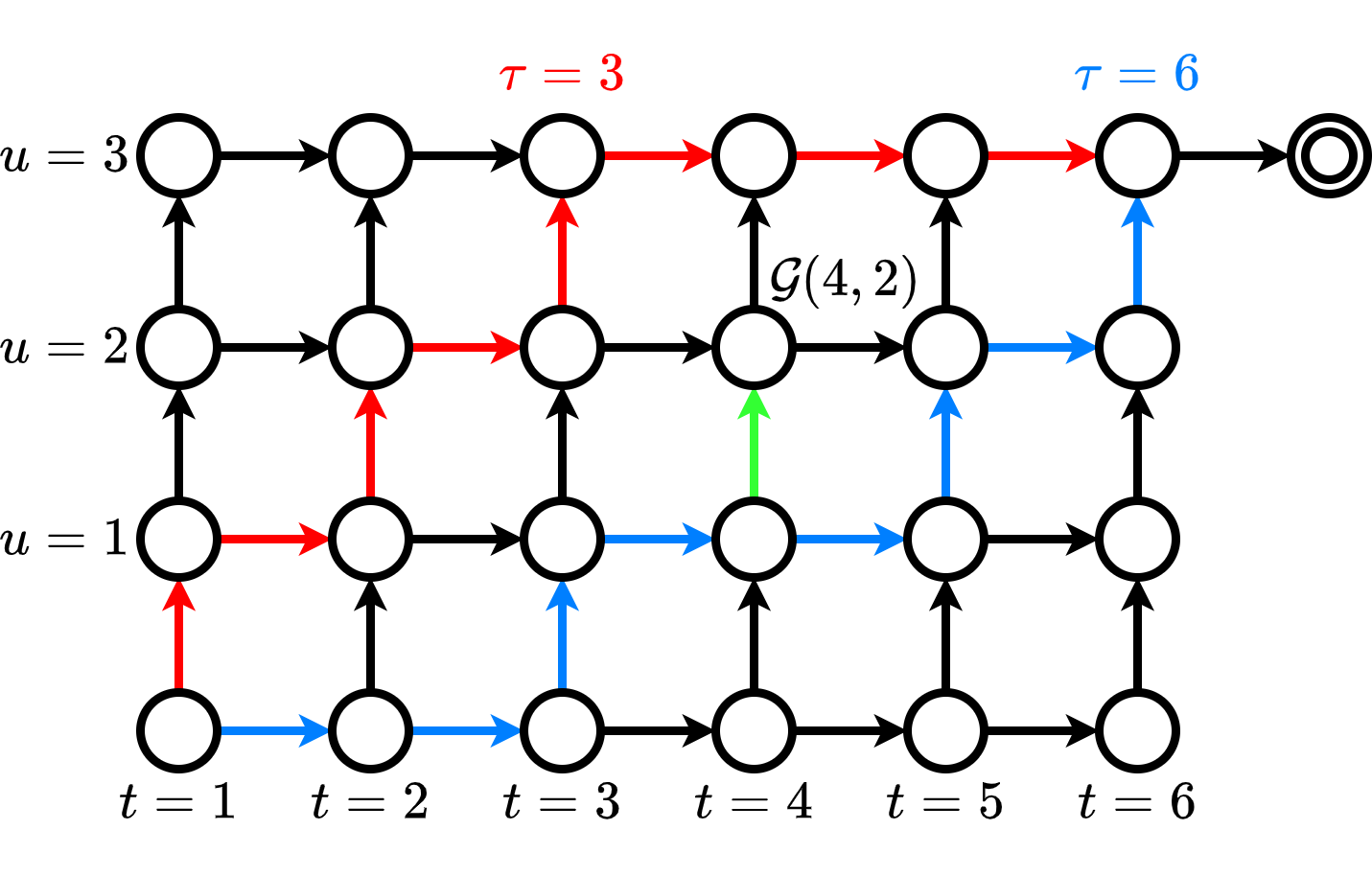} \\ (a) Transducer Lattice
\end{minipage}
} \quad 
\subfigure{
\begin{minipage}[t]{0.27\linewidth}
\centering
\includegraphics[width=\linewidth]{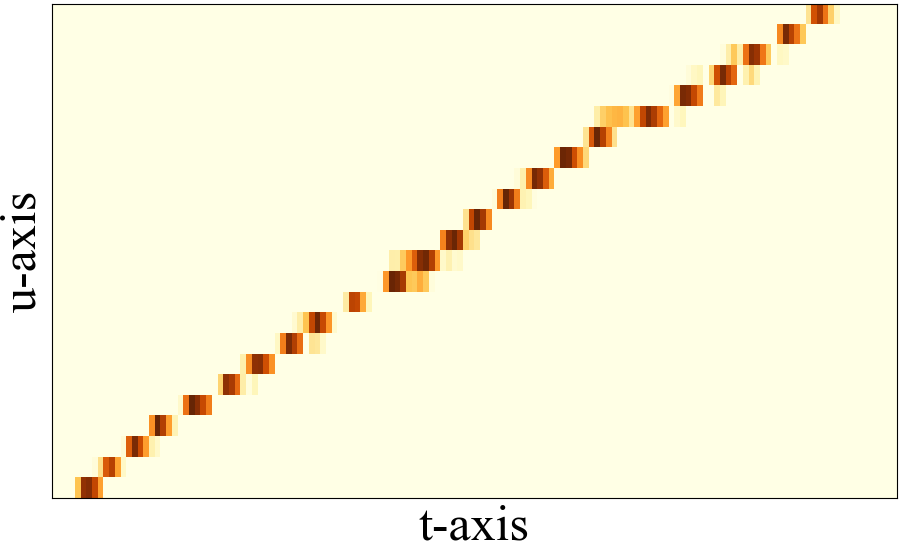} \\ (b) $\log \mathcal{G}(t,u)$ - Vanilla Transducer
\end{minipage}
} \quad \quad 
\subfigure{
\begin{minipage}[t]{0.30\linewidth}
\centering
\includegraphics[width=\linewidth]{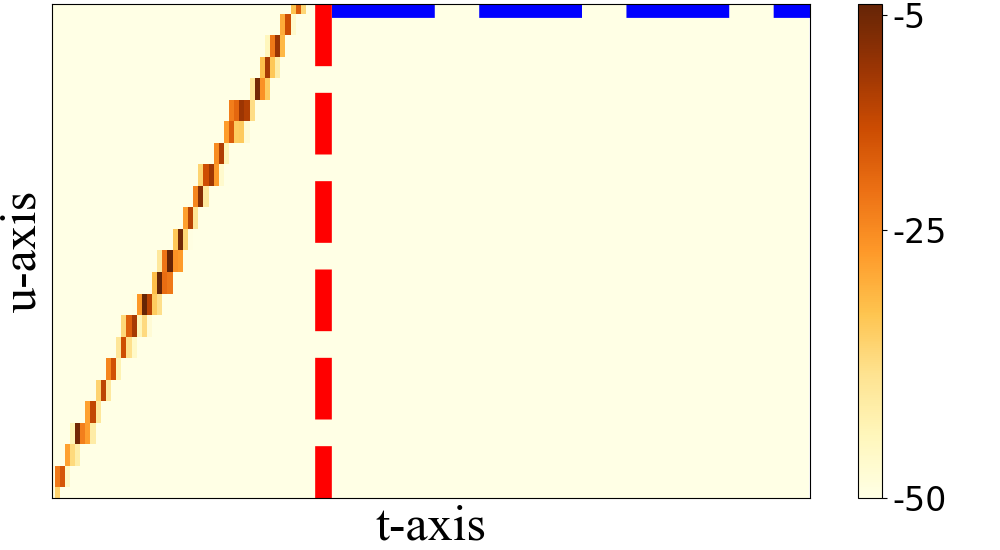} \\ (c) $\log \mathcal{G}(t,u)$ - BRT \quad \quad 
\end{minipage}
}
\caption{(a): Transducer lattice. $\mathcal{G}(4,2)$ is the summed posterior of all paths that go through the vertical arrow (in green) from node $(4,1)$ to node $(4,2)$ and emit 2nd token at the 4th frame. The red path ends all non-blank predictions at $\tau=3$ while the blue path ends at $\tau=6$. The red path is preferred in Sec.\ref{para_exp_offline}. (b) \& (c): The heat maps for $\log \mathcal{G}(t,u)$.} 
\label{fig_lattice}
\end{figure*}

\vspace{-5pt}
\subsection{Vanilla Transducer}
\vspace{-8pt}
In training, vanilla transducer maximizes the posterior of the transcription $\mathbf{l}=[l_1, ..., l_U]$ with the given acoustic feature sequence $\mathbf{x}=[\mathbf{x}_1, ..., \mathbf{x}_T]$.
Instead of maximizing $P(\mathbf{l}|\mathbf{x})$ directly, the transducer maximizes the summed posterior of all \textit{paths} in the transducer \textit{lattice} (see Fig.\ref{fig_lattice}.a). Note $\varnothing$ as the blank symbol and extend the vocabulary $\hat{\mathcal{V}}=\mathcal{V}\cup\{\varnothing\}$, each symbol sequence $\mathbf{\pi}=[\pi_1, ..., \pi_{T+U}]$ is a valid path if all entries of $\mathbf{\pi}$ are in $\hat{\mathcal{V}}$ and $\mathcal{B}(\mathbf{\pi})=\mathbf{l}$. Here $\mathcal{B}$ is a mapping that removes all $\varnothing$. So the vanilla transducer objective to minimize is defined as:
\begin{equation}
\setlength{\abovedisplayskip}{0pt}
\setlength{\belowdisplayskip}{0pt}
\label{def_nt}
   \mathbf{J}_{\text{transducer}}(\mathbf{l},\mathbf{x}) \triangleq -\log P(\mathbf{l}|\mathbf{x}) = -\log\sum_{\mathbf{\pi}\in\mathcal{B}^{-1}(\mathbf{l})} P(\mathbf{\mathbf{\pi}}|\mathbf{x})
\end{equation}
where $\mathcal{B}^{-1}(\mathbf{l})$ is the set of all valid paths. 
Next, the posterior of each path $P(\mathbf{\pi}|\mathbf{x})$ is computed as:
\begin{equation}
\label{token_posterior}
\setlength{\abovedisplayskip}{0pt}
\setlength{\belowdisplayskip}{0pt}
    P(\mathbf{\pi}|\mathbf{x})=\prod_{i=1}^{T+U}p(\pi_{i}|\mathbf{x}_{1:t}, \mathbf{l}_{1:u})
\end{equation}
where the condition $(\mathbf{x}_{1:t}, \mathbf{l}_{1:u})$ specifies the node $(t,u)$ on the transducer lattice s.t. $\mathcal{B}(\mathbf{\pi}_{1:i})=\mathbf{l}_{1:u}$ and $t+u=i-1$.
Instead of enumerating all paths and summing their posteriors, the transducer objective is computed efficiently by \textit{forward-backward algorithm} \cite{fb_algo}, which recursively computes the forward-backward variables $\alpha(t,u)$ and $\beta(t,u)$ for each node $(t,u)$ in the transducer lattice:
\begin{equation}
\label{alpha}
\setlength{\abovedisplayskip}{0pt}
\setlength{\belowdisplayskip}{0pt}
    \alpha(t,u) = \sum_{\mathbf{\pi}\in \mathcal{V'}^{(T+U)}, \mathcal{B}(\mathbf{\pi}_{1:t+u})=\mathbf{l}_{1:u} } P(\mathbf{\pi}|\mathbf{x})
\end{equation}
\begin{equation}
\label{beta}
\setlength{\abovedisplayskip}{0pt}
\setlength{\belowdisplayskip}{0pt}
    \beta(t,u) = \sum_{\mathbf{\pi}\in \mathcal{V'}^{(T+U)}, \mathcal{B}(\mathbf{\pi}_{t+u+1:T+U})=\mathbf{l}_{u+1:U}} P(\mathbf{\pi}|\mathbf{x})
\end{equation}
Subsequently, by decomposing each path $\mathbf{\pi}$ into partial paths $\mathbf{\pi}_{1:t+u}$ and $\mathbf{\pi}_{t+u+1:T+U}$ and using Eq.\{\ref{alpha}, \ref{beta}\}, the transducer objective is derived as\footnote{Note, for a path that go through a known node $(t,u)$ in transducer lattice, the partial path $\mathbf{\pi}_{t+u+1:T+U}$ is independent to partial path $\mathbf{\pi}_{1:t+u}$ so that the factorization of the path posterior is $ P(\mathbf{\pi}|\mathbf{x}) = P(\mathbf{\pi}_{1:t+u}|\mathbf{x}) \cdot P(\mathbf{\pi}_{t+u+1:T+U}|{\cancel{\mathbf{\pi}_{1:t+u}}}, \mathbf{x})$.}:
\begin{align}
\setlength{\abovedisplayskip}{0pt}
\setlength{\belowdisplayskip}{0pt}
\label{def_occu}
    \mathbf{J}_{\text{transducer}}(\mathbf{l},\mathbf{x})
    &= -\log \sum_{\mathbf{\pi}\in\mathcal{B}^{-1}(\mathbf{l})} P(\mathbf{\mathbf{\pi}}|\mathbf{x}) \notag \\
    =& - \log \sum_{(t,u): t+u=n} \sum_{\mathcal{B}(\mathbf{\pi}_{1:t+u})=\mathbf{l}_{1:u} \atop \mathcal{B}(\mathbf{\pi}_{t+u+1:T+U})=\mathbf{l}_{u+1:U}} P(\mathbf{\mathbf{\pi}}|\mathbf{x}) \notag \\ 
    =& - \log \sum_{(t,u):t+u=n}\alpha(t,u)\cdot\beta(t,u)
\end{align}
where $n$ is any known integer s.t. $n\in[0, T+U]$.
Finally, the path with the highest posterior is usually considered as the alignment prediction between $\mathbf{x}$ and $\mathbf{l}$: $\text{ali}(\textbf{x},\textbf{l}) = \arg\max_{\mathbf{\pi\in\mathcal{B}^{-1}(\mathbf{l})}}P(\mathbf{\pi}|\mathbf{x})$.

\begin{figure*}[t]
\setlength{\belowcaptionskip}{-0.5cm}
\setlength{\abovecaptionskip}{-0.0cm}
\centering
\subfigure{
\begin{minipage}[t]{0.305\linewidth}
\centering
\includegraphics[width=\linewidth]{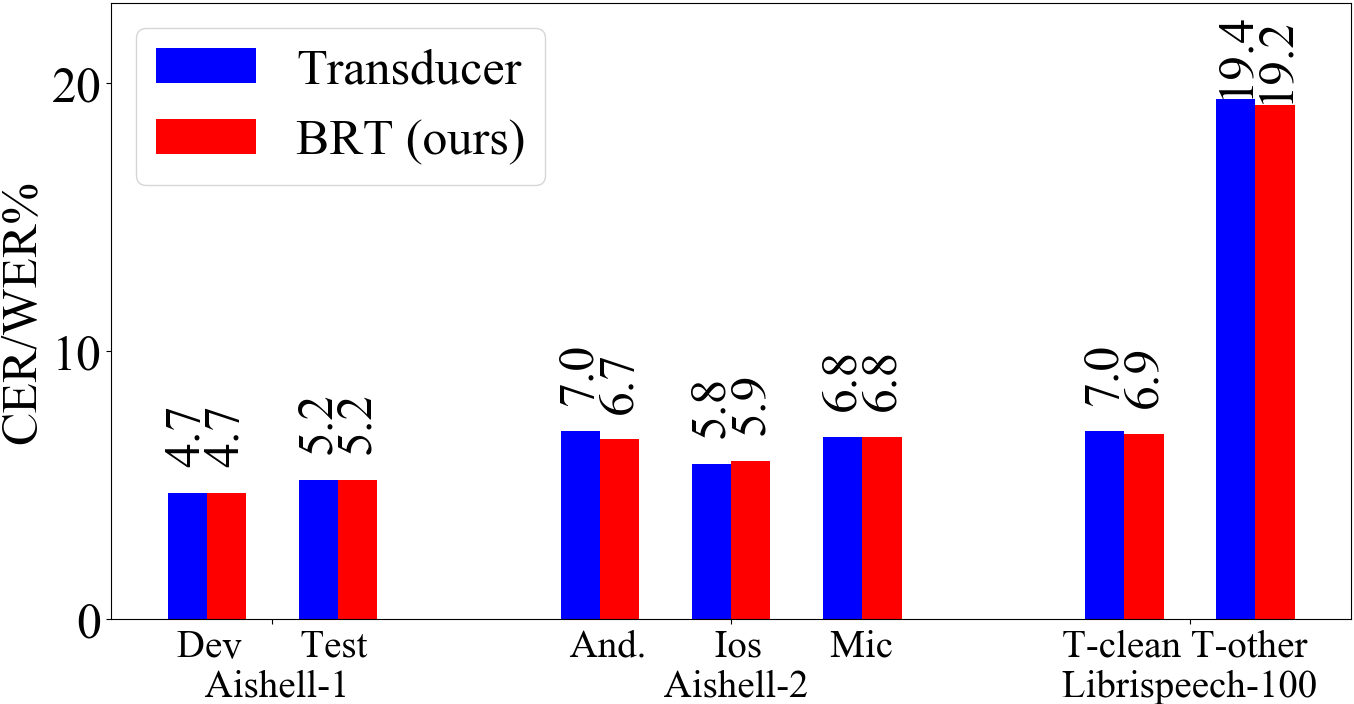} \\ (a) WER / CER\%
\end{minipage}
} \ \ 
\subfigure{
\begin{minipage}[t]{0.315\linewidth}
\centering
\includegraphics[width=\linewidth]{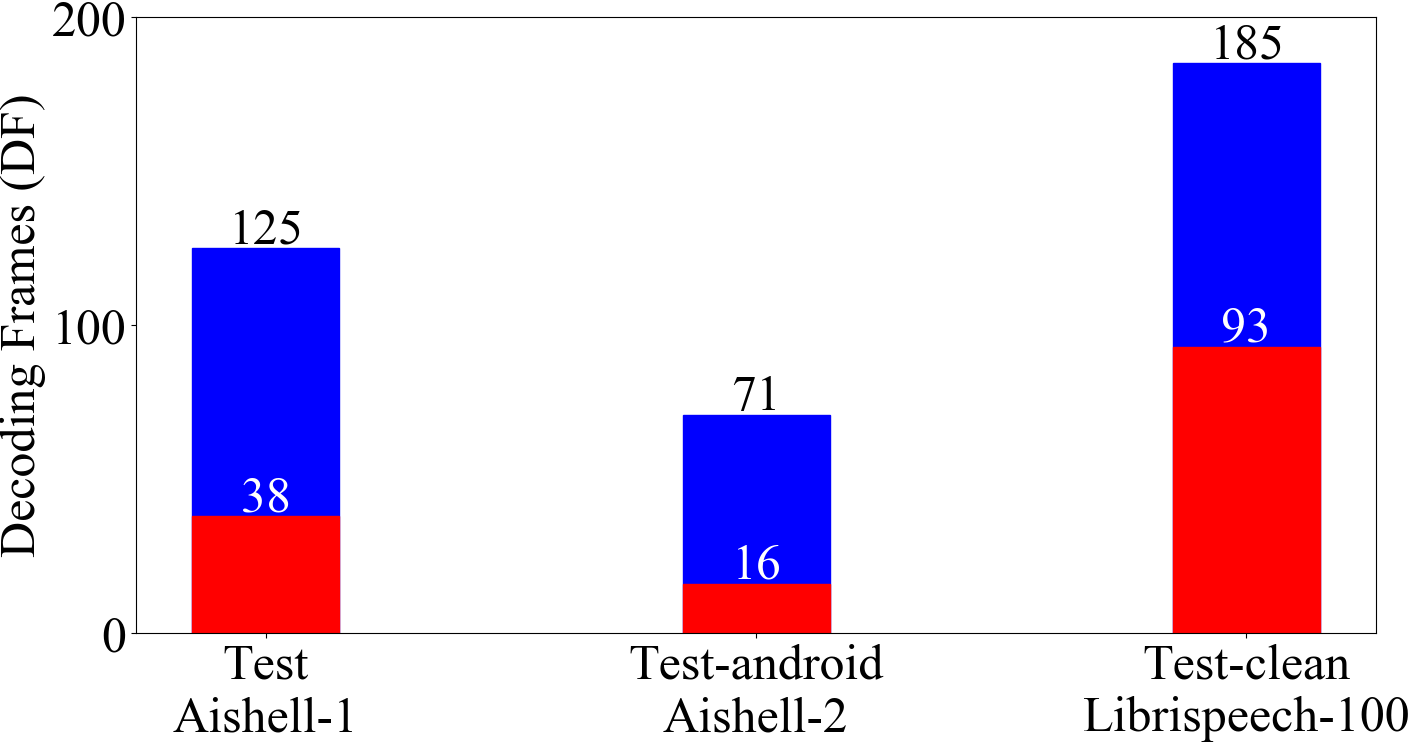} \\ (b) Decoding Frames (DF)
\end{minipage}
} \ \ 
\subfigure{
\begin{minipage}[t]{0.305\linewidth}
\centering
\includegraphics[width=\linewidth]{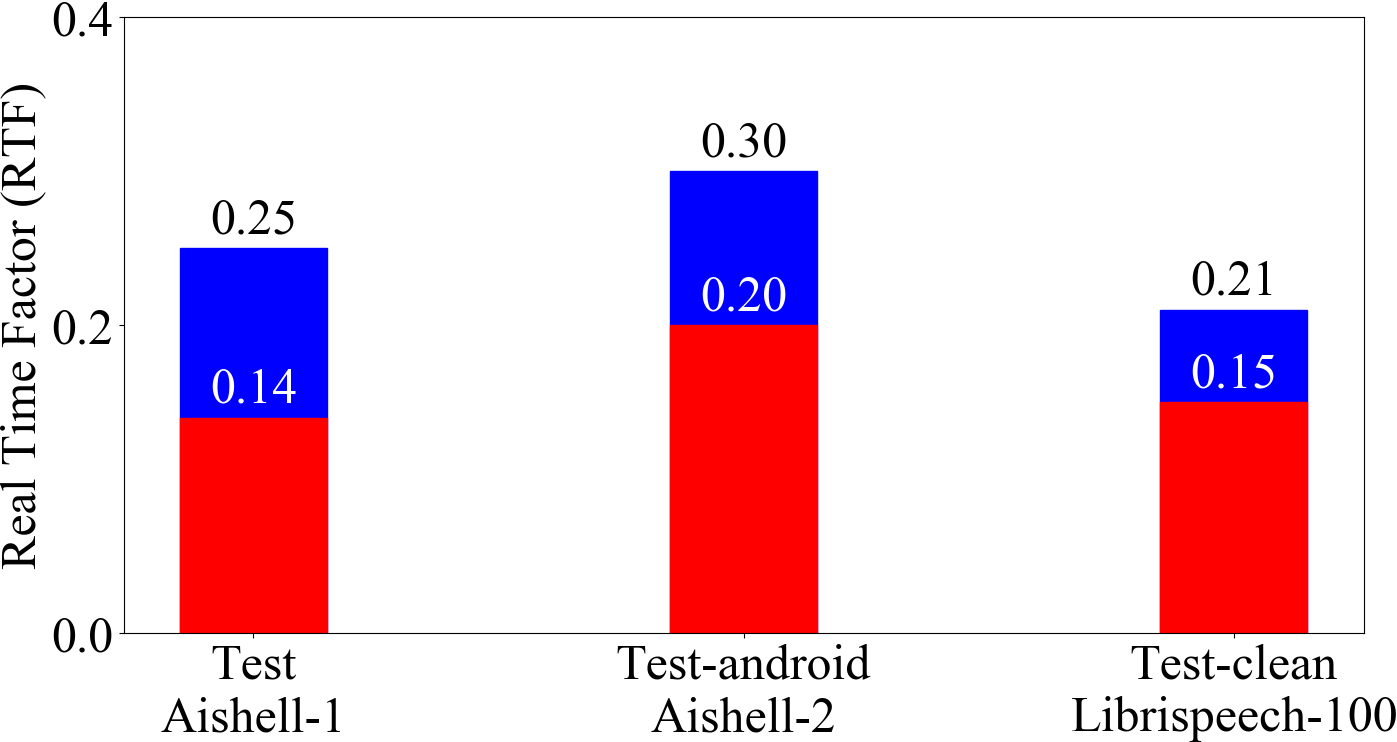} \\ (c) Real Time Factor (RTF)
\end{minipage}
}
\caption{Results of non-streaming ASR. (a) Recognition accuracy in CER / WER \%; (b) Average decoding frames (DF); (c) Real time factor (RTF). With comparable recognition accuracy, the proposed BRT achieve more efficient decoding by decoding fewer frames.}
\label{fig_exp_offline}
\end{figure*}

\vspace{-5pt}
\subsection{Bayes Risk Transducer}
\label{para_brt}
\vspace{-5pt}
As suggested in Eq.\ref{def_nt}, the formulation of the vanilla transducer has no prior preference among paths. This work intentionally selects the predicted alignment among the paths and thus attempts to achieve controllable alignment prediction. With this motivation, this work proposes Bayes Risk Transducer (BRT), which adopts a customizable Bayes risk function to express the preference for specific paths with desired properties. 
To preserve a similar format like Eq.\ref{def_nt}, we sets the risk function for each path as $-r(\mathbf{\pi})$ so that minimizing the expected risk is equivalent to minimizing the BRT objective
\footnote{In the remained part of this work, $r(\mathbf{\pi})$ is termed as the Bayes risk function even though the real Bayes risk function is $-r(\mathbf{\pi})$. Higher $r(\mathbf{\pi})$ value represents a lower risk, a.k.a., preference.}:
\begin{equation}
\setlength{\abovedisplayskip}{0pt}
\setlength{\belowdisplayskip}{0pt}
\label{def_brt}
   \mathbf{J}_{\text{BRT}}(\mathbf{l},\mathbf{x}) \triangleq -\log \sum_{\mathbf{\pi}\in\mathcal{B}^{-1}(\mathbf{l})} [P(\mathbf{\mathbf{\pi}}|\mathbf{x})\cdot r(\mathbf{\pi})]
\end{equation}

The computation of the proposed BRT objective still adopts the forward-backward variables and works in a divide-and-conquer approach. To express the preference for specific property of the paths, the Bayes risk function specifies 1) what property is concerned and 2) what values of the concerned property are preferred. To answer these questions, the concerned property of each path is defined by $f(\mathbf{\pi})$. Then, all paths are divided into multiple exclusive groups s.t. paths with the identical concerned property value $\tau$ are in the same groups. For paths in one identical group, it is reasonable to assign the same risk value as the concerned property is the same. Thus, the Bayes risk function $r(\mathbf{\pi})$ is replaced by a group-level risk function $r_g(\tau)$, which only depends on the group-level concerned property $\tau$ rather than the path $\mathbf{\pi}$. Formally, this process is written as:
\begin{align}
\setlength{\abovedisplayskip}{-5pt}
\setlength{\belowdisplayskip}{-5pt}
\label{eq_group}
    \mathbf{J}_{\text{BRT}}(\mathbf{l},\mathbf{x}) 
    & =  -\log \sum_{\tau}\sum_{\mathbf{\pi}\in\mathcal{B}^{-1}(\mathbf{l}), f(\mathbf{\pi})=\tau} [P(\mathbf{\mathbf{\pi}}|\mathbf{x})\cdot r(\mathbf{\pi})]  \notag \\
    & = -\log  \sum_{\tau}\sum_{\mathbf{\pi}\in\mathcal{B}^{-1}(\mathbf{l}), f(\mathbf{\pi})=\tau} [P(\mathbf{\mathbf{\pi}}|\mathbf{x})\cdot r_g(\mathbf{\tau})] \notag \\
    & =  -\log \sum_{\tau} [r_g(\tau) \cdot \sum_{\mathbf{\pi}\in\mathcal{B}^{-1}(\mathbf{l}), f(\mathbf{\pi})=\tau} P(\mathbf{\mathbf{\pi}}|\mathbf{x})] 
\end{align}

Please be aware of: 
1) when splitting the path into groups, the groups are supposed to be mutually exclusive so that each path is considered once and only once; 
2) by adopting the group-level risk function, we can avoid the complex weighted summation over all paths within each group.
3) the summed posterior of each path group should be fully tractable by the forward-backward variables so the computation remains efficient; 
4) by pursuing the desired properties in the predicted alignment between $\mathbf{x}$ and $\mathbf{l}$ during training, these properties are expected to be preserved in the predicted alignments between the test speech and the textual hypotheses during decoding.

Provided the general formulation of BRT in Eq.\ref{eq_group}, a naive example is in Eq.\ref{def_occu}, where the concerned property $\tau$ represents the pair $(t,u)$ and $r_g(\tau)=1$. Under this setting, the vanilla transducer is a special case of the proposed BRT.
Alternatively, given the $u$-th non-blank token $l_u$ in the transcription, another useful example is to set the concerned property $\tau$ as the time stamp when $l_u$ is emitted, a.k.a, $\pi_{\tau + u}=l_u$.
With a similar factorization like Eq.\ref{def_occu} and considering Eq.\{\ref{token_posterior}, \ref{alpha}, \ref{beta}\}, the BRT objective is further revised as:
\begin{align}
\setlength{\abovedisplayskip}{-5pt}
\setlength{\belowdisplayskip}{-5pt}
\label{def_ending}
     & \mathbf{J}_{\text{BRT}} (\mathbf{l},\mathbf{x}, u)
     =  -\log  \sum_{\tau} [r_g(\tau) \cdot \sum_{\mathbf{\pi}\in\mathcal{B}^{-1}(\mathbf{l}), \pi_{\tau+u}=l_u} P(\mathbf{\mathbf{\pi}}|\mathbf{x})] \notag \\
    & = - \log  \sum_{\tau} [r_g(\tau) \cdot \sum_{\substack{\mathcal{B}(\mathbf{\pi}_{1:\tau+u-1})=\mathbf{l}_{1:u-1} \\ 
    \pi_{\tau+u}=l_u \\ 
    \mathcal{B}(\mathbf{\pi}_{\tau+u+1:T+U})=\mathbf{l}_{u+1:U}}} 
    P(\mathbf{\mathbf{\pi}}|\mathbf{x})] \notag \\
    & = - \log \sum_{\tau} r_g(\tau) \cdot \underbrace{[\alpha(\tau, u-1) \cdot p(l_u|\mathbf{x}_{1:\tau}, \mathbf{l}_{1:u-1}) \cdot \beta(\tau, u)]}_{\triangleq \ \  \mathcal{G}(\tau, u)}
\end{align}
Here is $\mathcal{G}(\tau, u)$ the summed posterior of all paths that go through the vertical arrow from node $(\tau, u-1)$ to node $(\tau,u)$. 
Fig.\ref{fig_lattice}.a gives a demonstration of $\mathcal{G}(\tau, u)$ in the lattice and Fig.\ref{fig_lattice}.\{b, c\} provides numerical examples of $\mathcal{G}(\tau, u)$.
$\mathcal{G}(\tau, u)$ measures the summed probability of all valid paths that $l_u$ is emitted at $\tau$-th frame, which indicates the alignment prediction.
So far the group-level risk function $r_g(\tau)$ is not defined. Below we show two applications of Eq.\ref{def_ending} with different $r_g(\tau)$ designs.

\vspace{-5pt}
\subsection{Non-streaming Application: Efficient Decoding}
\vspace{-5pt}
\label{para_offline}
For frame-synchronized decoding algorithms of the transducer \cite{rnnt, tsd}, the inference cost highly depends on $T$ as the decoding loop is conducted frame-by-frame.
In Fig.\ref{fig_lattice}.b, the whole sequence $\mathbf{l}$ cannot be predicted until all frames are explored.
By contrast, in Fig.\ref{fig_lattice}.c, all non-blank tokens are emitted before reaching the red line, which allows us to stop decoding at an early time stamp (e.g., the red line) to save computation.

To achieve the heat map like Fig.\ref{fig_lattice}.c, the concerned property is exactly the time stamp $\tau$ when the last token $l_U$, as well as the whole sequence, is emitted: $\pi_{\tau+U}=l_U$. In addition, paths with smaller $\tau$ are preferred (see Fig.\ref{fig_lattice}.a) since fewer frames are consumed to predict all tokens. Set $u=U$ in Eq.\ref{def_ending}, the objective to minimize is: 
\begin{equation}
\setlength{\abovedisplayskip}{0pt}
\setlength{\belowdisplayskip}{-2pt}
\label{def_offline}
     \mathbf{J}_{\text{BRT}} (\mathbf{l},\mathbf{x}, U) =  - \log \sum_{\tau}\text{min}(e^{-\lambda \cdot (\tau-m\cdot U)/T}, 1) \cdot \mathcal{G}(\tau, U)
\end{equation}
where the risk function $r_g(\tau) = \text{min}(e^{-\lambda \cdot (\tau-m\cdot U)/T}, 1)$ expresses the preference for $\tau\in[1, m\cdot U]$ and shows exponentially decayed interest in $\tau > m \cdot U$\footnote{We do not express an extra preference for very small $\tau$ so it is less likely that multiple non-blank tokens are emitted at a single frame.}.
$\lambda$ and $m$ are hyper-parameters. $m$ is empirically set to 2 and $\lambda$ varies according to datasets.

This work further provides an early-stop mechanism to reduce the number of decoding frames of BRT.
First, assume we obtain a hypothesis $\mathbf{\hat{l}}=[\hat{l}_1, ..., \hat{l}_u]$ at $\tau$-th frame during decoding. The hypothesis is considered complete if no additional non-blank tokens are expected to be emitted in the search process over the remaining frames after $\tau$. 
In other words, for any possible path of the complete $\mathbf{\hat{l}}$, its sub-path after the $\tau$-th frame only consists of continuous $\varnothing$ (see the blue line in Fig.\ref{fig_lattice}.c).
So $\mathbf{\hat{l}}$ is considered complete only if the accumulated probability of the continuous $\varnothing$ since the $\tau$-th frame are with high confidence: $\sum_{t=\tau}^{T}\log p(\varnothing|\mathbf{x}_{1:t},\mathbf{\hat{l}}_{1:u}) > D$, where $D=-10$ is a threshold value\footnote{The computation for this condition cannot be considered as a search over the frames after $\tau$ since the series 
$\{p(\varnothing|\mathbf{x}_{1:t},\mathbf{\hat{l}}_{1:u})\}$ can be computed in parallel fashion and requires no loop.}.
Secondly, for a search beam that contains multiple hypotheses, we terminate the search when 1) the top $k=3$ blank-free hypotheses do not change for $f=5$ frames and 2) all top $k=3$ hypotheses are considered complete.

\vspace{-5pt}
\subsection{Streaming Application: Early Emission}
\vspace{-5pt}
\label{para_stream}
A streaming ASR system is expected to emit each token accurately and timely. The accuracy and latency, however, usually form a trade-off: better recognition accuracy requires longer context, which results in higher latency. 
For streaming ASR, BRT is designed to encourage all tokens to emit at early time stamps, even at the cost of slight performance degradation.
By doing so, BRT achieves a better accuracy-latency trade-off than the vanilla transducer, which is further demonstrated in Sec.\ref{para_exp_online}.

The vanilla transducer only attempts to transcribe the speech correctly but poses no constraint on when tokens would be emitted. 
By contrast, the proposed BRT can reduce the latency by enforcing the paths that emit each token at a smaller time stamp. Formally, with  any non-blank token $l_u$ and the exponentially decayed risk function $r_g(\tau)=e^{-\lambda \cdot (\tau - \tau')/T}$, a BRT objective is derived from Eq.\ref{def_ending} with the goal of encouraging $l_u$ to be emitted earlier:
\begin{equation}
\label{def_online_u}
\setlength{\abovedisplayskip}{0pt}
\setlength{\belowdisplayskip}{0pt}
    \mathbf{J}_{\text{BRT}} (\mathbf{l},\mathbf{x}, u) = - \log \sum_{\tau} e^{-\lambda \cdot (\tau - \tau')/T} \cdot \mathcal{G}(\tau, u)
\end{equation}
where $\tau$ is the concerned property that specifies the time-stamp when $l_u$ is emitted and is enforced to be smaller; $\tau'=\arg\max_{\tau}\mathcal{G}(\tau, u)$ is a bias term to ensure that the path group with the highest summed posterior $\mathcal{G}(\tau, u)$ would always receive the risk value of $r_g(\tau)=1$ so that the absolute value of $\mathbf{J}_{\text{BRT}} (\mathbf{l},\mathbf{x}, u)$ does not vary along with $u$ significantly. Here $\lambda$ is still an adjustable hyper-parameter varying with datasets.
Subsequently, to guide every token $l_u$ to be emitted earlier requires the consideration of all tokens. So we simply attempts to minimize the mean of $\mathbf{J}_{\text{BRT}} (\mathbf{l},\mathbf{x}, u)$ in Eq.\ref{def_online_u} over every $u$:
\begin{equation}
\setlength{\abovedisplayskip}{0pt}
\setlength{\belowdisplayskip}{0pt}
    \mathbf{J} (\mathbf{l},\mathbf{x}) =  \frac{1}{U} \cdot \sum_{u=1}^{U} \mathbf{J}_{\text{BRT}} (\mathbf{l},\mathbf{x}, u)
\end{equation}


\vspace{-5pt}
\section{Experiments}
\vspace{-5pt}
\setlength{\abovedisplayskip}{-0cm}
\setlength{\belowdisplayskip}{-0cm}
\begin{figure*}[t]
\centering
\subfigure{
\begin{minipage}[t]{0.33\linewidth}
\centering
\includegraphics[width=\linewidth]{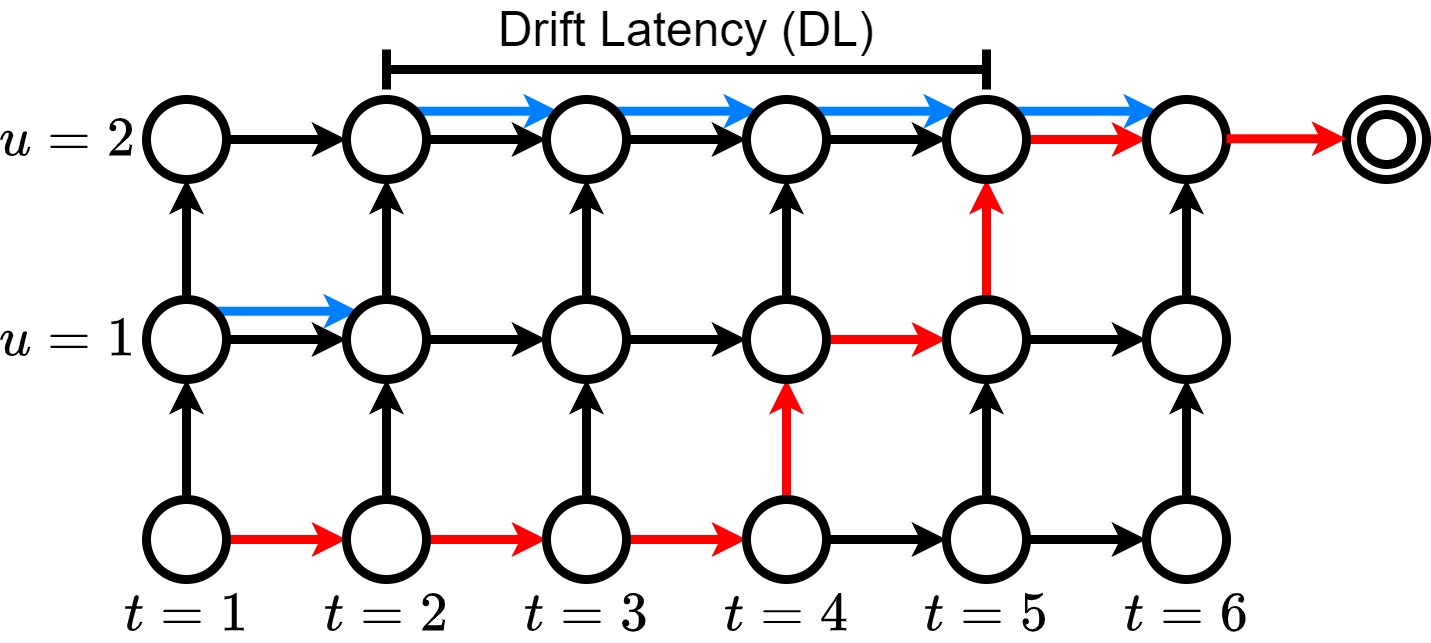} \\ (a) Demonstration of drift latency
\end{minipage}
}
\subfigure{
\begin{minipage}[t]{0.205\linewidth}
\centering
\includegraphics[width=\linewidth]{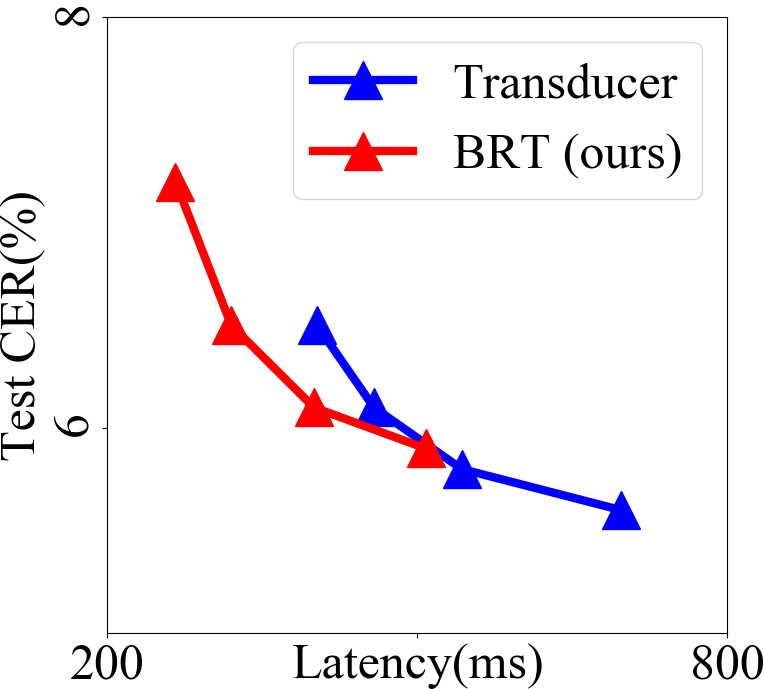} \\ (b) Aishell-1
\end{minipage}
}
\subfigure{
\begin{minipage}[t]{0.205\linewidth}
\centering
\includegraphics[width=\linewidth]{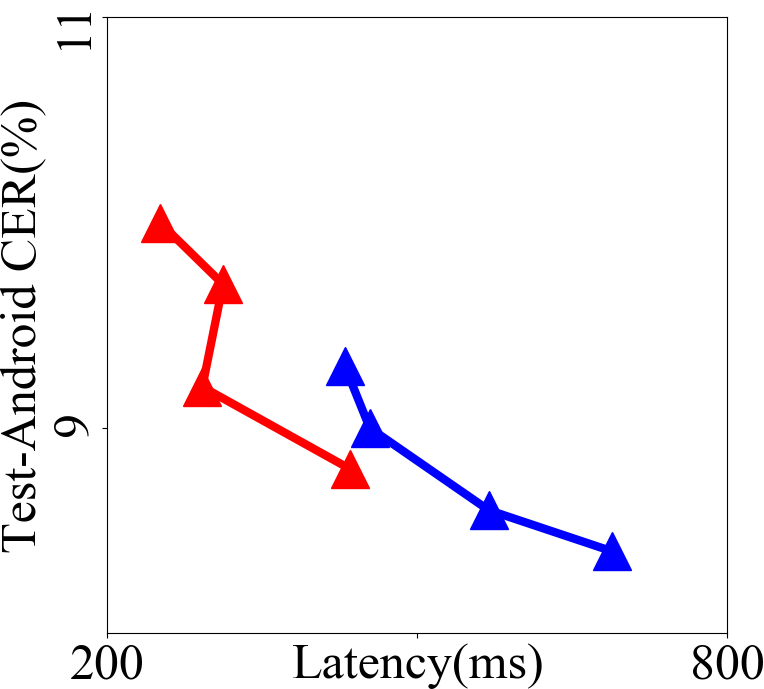} \\ (c) Aishell-2
\end{minipage}
}
\subfigure{
\begin{minipage}[t]{0.205\linewidth}
\centering
\includegraphics[width=\linewidth]{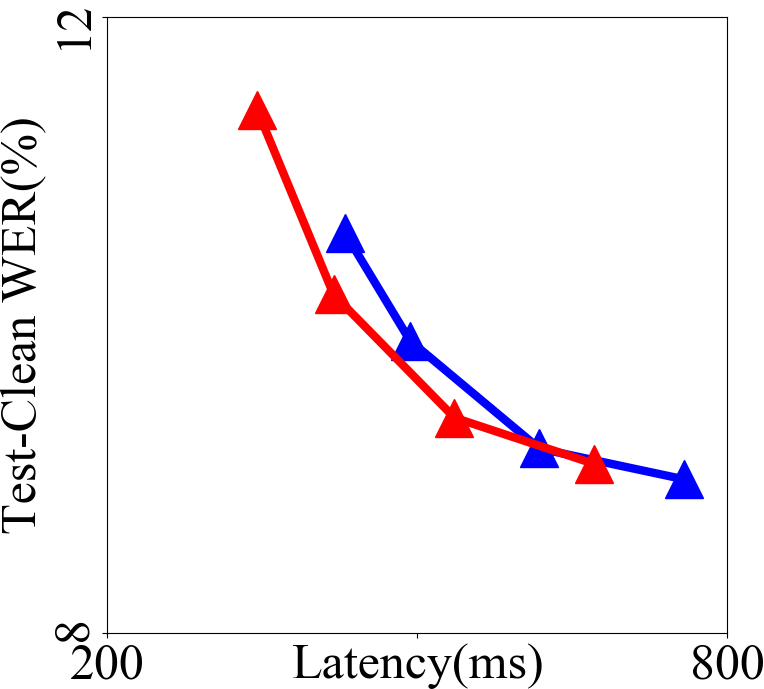} \\ (d) Librispeech-100
\end{minipage}
}
\caption{Results of streaming ASR. (a) A demonstration of drift latency (DL). Blue arrows stand for the reference duration of each token. The alignment of the hypothesis is represented by the red path. Token $l_2$ starts at 1st frame but is predicted at 4th frame, which is a 3-frame drift latency. (b) \& (c) \& (d): the accuracy-latency trade-off achieved with varying data collecting latency (DCL)}.
\label{fig_exp_online}
\end{figure*}
\subsection{Experimental Setup}
\vspace{-5pt}
\textbf{Datasets:} Experiments are conducted on Aishell-1 \cite{aishell1}, Aishell-2 \cite{aishell2} and Librispeech-100 \cite{librispeech} datasets. The volumes of these datasets range from 100 hours to 1k hours. Librispeech-100 is in English and the others are in Mandarin. All data is augmented by SpecAugment \cite{specaug} and speed perturbation. For English, tokens are 500 BPE units. \\
\textbf{Evaluation Metrics:} CER / WER\% is adopted to show the recognition accuracy. To compare the decoding efficiency of non-streaming ASR, the average number of decoding frames (DF) before the decoding termination and the real-time factor (RTF) over CPU\footnote{Intel(R) Xeon(R) Platinum 8255C CPU @ 2.50GHz, single thread.} are reported. For streaming ASR, the overall latency is defined as the sum of data collecting latency (DCL) and drift latency (DL) \cite{brctc}\footnote{The latency caused by computation is not considered in this work: it is marginal for a light model and without an external language model.}. DCL is the time to wait before the input speech forms a chunk (a.k.a., the latency caused by chunk size and look-ahead length). DL is exemplified in Fig.\ref{fig_exp_online}.a. 
and its reference\footnote{For an English word that consists of multiple BPE tokens, we only count the last BPE unit of that word.} is obtained by standard GMM-HMM systems\footnote{Models are trained by Kaldi: https://github.com/kaldi-asr/kaldi}.\\
\textbf{Features \& Models:} 80-dim Fbank features with the window size of 10ms are down-sampled by 4x using CNN before being fed into the encoder. The acoustic encoder is Conformer \cite{conformer} for non-streaming ASR and Emformer \cite{emformer} for streaming ASR. For streaming ASR, DCL is set to \{160, 320, 480, 640\}ms. The prediction network is a standard LSTM and the joint network is linear. For English tasks, an auxiliary CTC criterion is adopted on the top of the encoder to stabilize the training. Model sizes for non-streaming and streaming experiments are 95M and 57M respectively.\\
\textbf{Training \& Decoding:} For \{Aishell-1, Aishell-2, Librispeech-100\}, models are trained for \{100, 100, 300\} epochs with $\lambda$ of \{5, 5, 20\} and \{10, 10, 50\} for non-streaming and streaming ASR respectively. The original decoding algorithm proposed in \cite{rnnt} is adopted with a beam size of 10. No language model is adopted in decoding. \\
\vspace{-15pt}
\subsection{Results on non-streaming ASR}
\vspace{-5pt}
\label{para_exp_offline}
This part evaluates the effectiveness of the proposed BRT method on non-streaming ASR. Our results are shown in Fig.\ref{fig_exp_offline}. Firstly, as shown in Fig.\ref{fig_exp_offline}.a, with datasets in varying scales and languages, the recognition accuracy achieved by the proposed BRT method and the vanilla transducer are comparable. Secondly, Fig.\ref{fig_exp_offline}.b demonstrates the effectiveness of the proposed BRT in reducing the decoding frames. E.g., by introducing the BRT criterion, the DF for Aishell-2 dataset is reduced from 71 to 16, which is a 77\% reduction. Finally, for models trained by BRT, the overall inference cost (a.k.a., RTF) is reduced since the proposed early-stop mechanism allows the model not to explore all the frames. By adopting the BRT criterion and the early-stop mechanism, the RTF of Aishell-1 is reduced from 0.25 to 0.14, which is a 46\% relative reduction. The reduction in Fig.\ref{fig_exp_offline}.c is not as considerable as that in Fig.\ref{fig_exp_offline}.b since the encoder inference accounts for a large part of the computation cost.

\vspace{-5pt}
\subsection{Results on streaming ASR}
\vspace{-5pt}
\label{para_exp_online}
This part evaluates the effectiveness of the proposed BRT method on streaming ASR. Our results are shown in Fig.\ref{fig_exp_online}. As discussed in Sec.\ref{para_stream}, the streaming ASR has a trade-off between the recognition accuracy and the overall system latency.
As shown in Fig.\ref{fig_exp_online}.\{b,c,d\}, on the three datasets, the curve of the proposed BRT (the red one) consistently lies in the lower-left direction of its baseline (vanilla transducer, the blue one), which suggested that the proposed BRT criterion achieves better accuracy-latency trade-off than vanilla transducer. In addition, BRT can build systems with extremely low latency that cannot be achieved by the vanilla transducer, even at the cost of recognition performance degradation. E.g., on the Aishell-2 dataset, the lowest overall latency achieved by the vanilla transducer and BRT is 430ms and 251ms respectively, which is a 41\% relative reduction in latency, even with accuracy degradation.

\begin{figure}[h]
\setlength{\abovedisplayskip}{-1pt}
\setlength{\belowdisplayskip}{-1pt}
\centering
\subfigure{
\begin{minipage}[t]{0.45\linewidth}
\centering
\includegraphics[width=\linewidth]{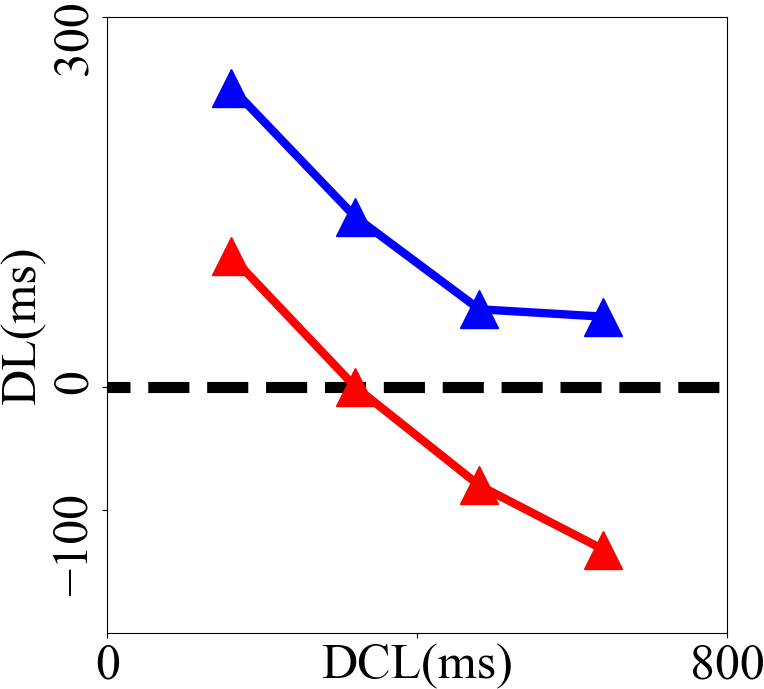} \\ (a) DCL vs. DL
\end{minipage}
}
\subfigure{
\begin{minipage}[t]{0.45\linewidth}
\centering
\includegraphics[width=\linewidth]{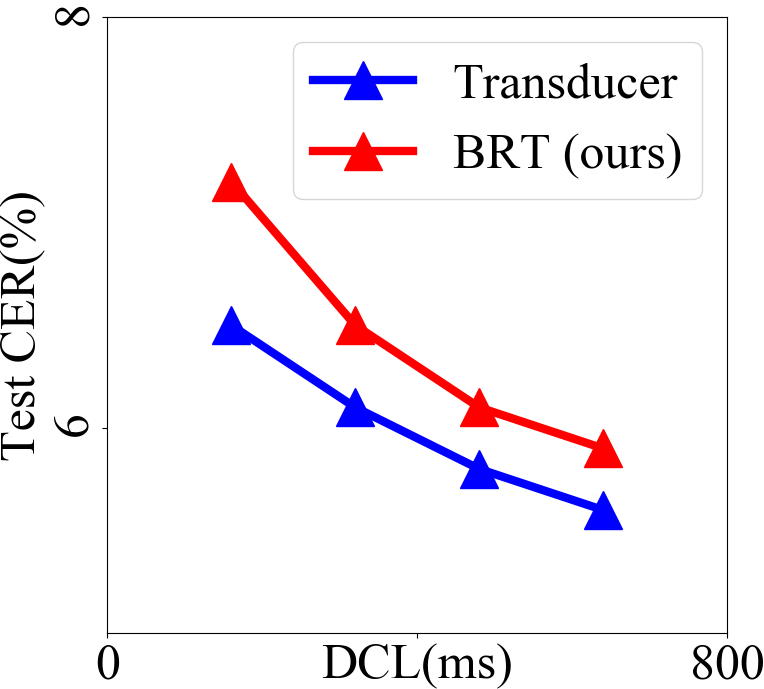} \\ (b) DCL vs. CER
\end{minipage}
}
\caption{Ablation study for streaming ASR (on Aishell-1)}
\label{fig_exp_abla}
\end{figure}

Further ablation study is conducted on Aishell-1 dataset. As shown in Fig.\ref{fig_exp_abla}.a, the transducer system with extremely low latency cannot be built by simply reducing the chunk size (a.k.a., small DCL) since the model is allowed to wait for very long context before emitting (a.k.a., larger DL). In addition, the adoption of BRT can effectively reduce the DL, which is aligned with our motivation in Sec.\ref{para_stream}\footnote{The DL can be negative due to the look-ahead of the model}. The BRT model has this strength of early emission since the paths that emit non-blank prediction earlier are enforced during training.
Next, Fig.\ref{fig_exp_abla}.b shows that the vanilla transducer outperforms the proposed BRT in accuracy with all DCL settings, which is reasonable since the accessible right context is reduced if the non-blank tokens are emitted earlier. 
Combining Fig.\ref{fig_exp_abla}.\{a,b\} will reach Fig.\ref{fig_exp_online}.b, which demonstrates that: BRT provides an alternative solution for streaming transducer, i.e., increasing DCL with a larger chunk size and reducing DL by using BRT to meet the latency budget so that a better overall accuracy-latency trade-off is achieved.
\vspace{-10pt}

\vspace{-5pt}
\section{Conclusion}
\vspace{-5pt}
To achieve controllable alignment prediction in the transducer, this work proposes an extension of the transducer called Bayes Risk Transducer (BRT), which adopts a Bayes risk function to enforce specific paths with the desired properties.
By designing different Bayes risk functions, the predicted alignment is enriched with task-specific properties, which provides practical benefits besides recognizing the speech accurately: efficient decoding for non-streaming ASR and early emission for streaming ASR. The claimed two applications are experimentally validated on multiple datasets and in multiple languages.

\bibliographystyle{IEEEtran}
\bibliography{mybib}

\end{document}